\documentclass[lettersize,journal]{IEEEtran}
\usepackage{amsmath}
\usepackage{amsmath,amsfonts}
\usepackage{amssymb}

\usepackage{multirow}
\usepackage{booktabs}
\usepackage{lipsum}
\usepackage{authblk}   % 必须加，才支持 \affiliation
\usepackage{adjustbox}
\usepackage[figuresright]{rotating}
\usepackage[ruled,linesnumbered]{algorithm2e}
\usepackage[colorlinks=true, allcolors=blue]{hyperref}

\usepackage[table]{xcolor}
\usepackage[caption=false,font=normalsize,labelfont=sf,textfont=sf]{subfig}
\usepackage{textcomp}
\usepackage{stfloats}
\usepackage{graphicx}
\usepackage{float}
\usepackage{cite}
\usepackage{marvosym}
\usepackage[pagewise]{lineno}
\author[1]{Yufei Zhang}
\author[1]{Chenlu Zhan}
\author[1]{Hongwei Wang*}

\affil[1]{Zhejiang University}

\begin{document}

\title{GaussianDS: Depth-supervised Semantic Gaussian Splatting for Scene Understanding}
\maketitle

\begin{abstract}
3D Gaussian Splatting provides an efficient representation for 3D reconstruction, and recent extensions attach semantic attributes to Gaussians for open-vocabulary scene understanding. However, lifting view-dependent 2D foundation-model outputs into 3D space introduces cross-view inconsistencies and weak geometric grounding, leading to severe semantic drift and boundary leakage. We propose GaussianDS, a depth-supervised semantic 3DGS framework that treats semantic lifting as a supervision-alignment problem and jointly optimizes RGB appearance, rendered depth, and compact semantics from scratch. Specifically, GaussianDS organizes unordered multi-view images into a pose-aware pseudo-video trajectory to propagate view-consistent masks via SAM2. During joint optimization, scale-shift-aligned monocular depth supervision and depth total-variation regularization stabilize Gaussian geometry, while a depth-edge-aware refinement loss explicitly anchors semantic transitions onto physical geometric discontinuities. Extensive evaluations show that our end-to-end framework not only retains high-fidelity 3D reconstruction and real-time rendering, but also establishes superior semantic understanding. GaussianDS sets new state-of-the-art performance on LERF (60.5\% mIoU) and 3D-OVS (97.79\% mIoU, 90.28\% mBIoU) by mitigating semantic leakage, while seamlessly facilitating downstream 3D object removal.
\end{abstract}


%==========================================================================
% SECTION I: INTRODUCTION
%==========================================================================
\section{Introduction}
\label{sec:intr}
As an explicit scene representation, 3D Gaussian Splatting (3DGS) \cite{3dgs_kerbl20233d} has emerged as an effective foundation for 3D reconstruction \cite{jin20243dfires,kerbl2024hierarchical}, dynamic rendering \cite{yu2024cogs}, object manipulation \cite{wang2024gscream}, and spatial segmentation \cite{gg_gaussian_grouping,zhou2024feature}. By parameterizing geometry and appearance with anisotropic Gaussian primitives optimized through differentiable splatting, 3DGS achieves high-fidelity real-time rendering. However, because its original formulation exclusively optimizes 3D reconstruction objectives, converting a purely visual appearance field into 3D semantic representation remains non-trivial.

\begin{figure}[ht]
    \centering
    \includegraphics[width=1\linewidth]{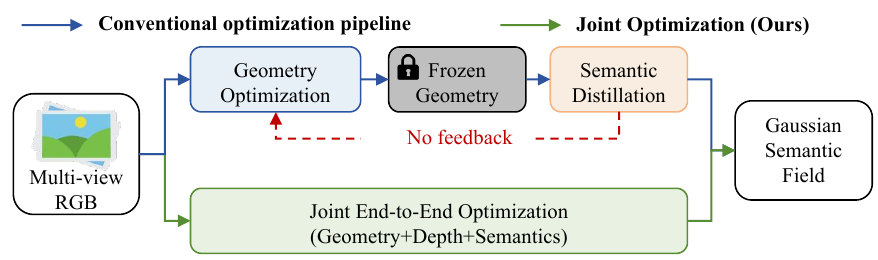}
    \caption{Conventional pipelines freeze geometry prior to semantic distillation (two stages), whereas GaussianDS performs joint end-to-end optimization of RGB, depth, and semantics.}
    \label{fig-motivation}
\end{figure}

To bypass expensive dense 3D annotations, recent semantic 3DGS methods lift 2D representations derived from vision-language models, such as CLIP~\cite{clip_radford2021learning}, SAM~\cite{sam_Kirillov_2023_ICCV}, and DINO~\cite{dino_Caron_2021_ICCV}, into 3D space. As shown Fig.\ref{fig-motivation}, conventional paradigms \cite{langsplat_qin2024langsplat,shi2024languageembedding} adopt a two-stage training pipeline that first pre-trains and freezes the geometric and photometric Gaussian field, subsequently distilling semantic features onto fixed primitives. While isolating semantic distillation simplifies optimization, this decoupled strategy precludes semantic feedback from regularizing Gaussian spatial distributions and remains fundamentally bounded by the cross-view consistency of the lifted 2D supervision.

We identify this bottleneck as a critical supervision-alignment deficiency. Predictions from 2D foundation models are intrinsically view-dependent and constrained to the image plane, frequently yielding discordant masks, ambiguous object contours, and fluctuating feature responses across varying viewpoints. Whereas 3D appearance reconstruction stably converges because color errors are continuous and directly governed by physical image formation, semantic supervision operates on discrete, boundary-sensitive object identities\cite{hu2026sagd}. When disparate 2D semantic targets are backpropagated into identical 3D Gaussians through visibility-based $\alpha$-compositing, conflicting gradients induce catastrophic semantic drift and boundary leakage across physical surfaces \cite{cen2025segment,zhu2025rethinking}.

Crucially, addressing this limitation requires geometric grounding rather than merely scaling semantic feature capacities. While structured semantic codebooks~\cite{opengaussian,tian2025ccllgs} and expressive latent embeddings~\cite{liao2024clipgs} enhance open-vocabulary feature retrieval, they fundamentally fail to eliminate the supervisory gradient conflicts induced by unaligned multi-view predictions. Geometrically, semantic boundaries must consistently coincide with physical depth discontinuities rather than arbitrary 2D segmentation errors. Readily accessible monocular depth estimators, such as Depth Anything\cite{depth_anything_yang2024depth} and ZoeDepth \cite{bhat2023zoedepth}, provide dense relative spatial arrangements and structural edge priors. Although monocular depth lacks absolute metric scale, its relative ordering and surface gradient constraints offer decisive guidance to regularize both Gaussian spatial layouts and semantic transitions.

Driven by these insights, we propose GaussianDS, a depth-supervised semantic 3DGS framework that unifies 2D semantic supervision and 3D Gaussian optimization. Unlike conventional cascaded pipelines that freeze geometry prior to semantic distillation, GaussianDS conducts joint end-to-end optimization of RGB appearance, rendered depth, and latent semantics from scratch. Before lifting features to 3D, GaussianDS organizes unordered multi-view captures into a pose-aware pseudo-video sequence, enabling SAM2 to propagate spatially coherent masklets across overlapping viewpoints and suppress inter-view mask jitter. During joint optimization, scale-shift-aligned monocular depth supervision and total-variation depth regularization constrain Gaussian geometry, while a depth-edge-aware refinement loss explicitly anchors semantic boundaries to physical geometric discontinuities. Mask-guided CLIP features are compressed into a compact latent space and jointly optimized with Gaussian geometric parameters, forming a unified, real-time renderable semantic Gaussian field. The primary contributions of this work are summarized as follows:
\begin{itemize}
\item{We formulate the supervision-alignment bottleneck in semantic 3DGS, demonstrating that unaligned 2D foundation-model predictions induce conflicting gradients on shared Gaussian primitives.}
\item{We develop GaussianDS, a joint optimization framework that simultaneously learns color, depth, and compact semantic fields from scratch, eliminating the need for cascaded two-stage pre-training pipelines.}
\item{We integrate pose-aware pseudo-video mask propagation with scale-shift depth alignment, rendered depth regularization, and depth-edge-aware semantic refinement to anchor semantic transitions onto scene geometry.}
\item{Extensive evaluations on LERF and 3D-OVS benchmarks show that GaussianDS achieves superior open-vocabulary query accuracy (60.5\% mIoU on LERF) and boundary localization (97.79\% mIoU and 90.28\% mBIoU on 3D-OVS) while maintaining real-time rendering.}
\end{itemize}

%==========================================================================
% SECTION II: RELATED WORK
%==========================================================================
\section{Related Work}
\label{sec:related}

\subsection{3D Gaussian Splatting and Geometric Regularization}
3D Gaussian Splatting (3DGS)~\cite{3dgs_kerbl20233d} parameterizes continuous 3D scenes via differentiable rasterization of anisotropic Gaussian primitives, achieving high-fidelity real-time novel view synthesis. To suppress geometric degradation such as floating artifacts and surface distortion, several works incorporate explicit surface and structural constraints. SuGaR~\cite{sugar_guedon2024sugar} binds flat Gaussians to implicit surface meshes through signed distance regularization, while NeuGS~\cite{chen2023neusg} and 3DGSR~\cite{lyu20243dgsr} derive surface normals directly from Gaussian density gradients to reconstruct planar regions with multi-view normal consistency. In parallel, external depth cues are widely integrated to resolve scale ambiguity and stabilize spatial distributions. DNGaussian~\cite{dngaussian} introduces a global-local depth normalization strategy to regularize depth gradients, Depth-Regularized 3DGS~\cite{depth_r_chung2024depth} optimizes relative depth rankings via Pearson correlation losses, and SDP-GS~\cite{zhao2026sdp} applies segmentation-aware depth filtering to prevent depth over-smoothing across distinct object boundaries. DGC-GS~\cite{zhang2026dgc} further reinforces geometric coherence by enforcing multi-view epipolar depth constraints under sparse camera configurations, while monocular depth optimization with dynamic $k$-nearest-neighbor (KNN) densification~\cite{wang20263d} ensures uniform primitive distribution across under-sampled surfaces. Although these geometric frameworks successfully stabilize spatial layouts, they restrict depth and geometric priors strictly to the appearance reconstruction pathway. In contrast, GaussianDS routes scale-shift aligned depth directly into the semantic optimization pipeline, explicitly utilizing physical depth discontinuities to regularize 3D semantic boundary transitions.

\subsection{2D Segmentation and Semantic Lifting}
The open-vocabulary capabilities of 2D vision-language and foundation models offer scalable supervisory signals for 3D spatial reasoning without requiring manual 3D annotations. Foundation models such as SAM \cite{sam_Kirillov_2023_ICCV}, LISA\cite{lai2024lisa}, and Grounded-SAM\cite{ren2024groundedsam} deliver robust zero-shot mask generation, while vision-language encoders like CLIP and DINO \cite{oquab2023dinov2} extract rich open-vocabulary semantics. To lift these 2D features into 3D space, SAI3D \cite{yin2024sai3d} aggregates SAM-derived region masks with multi-view CLIP embeddings through geometric consensus; LSeg \cite{lsegli2022languagedriven} enables language-driven pixel-level reasoning; and MPD-GS \cite{he2025mpd} leverages 2D boundary masks to guide adaptive Gaussian densification along object silhouettes. Nevertheless, 2D segmentation models process multi-view captures independently, inevitably causing semantic identity mismatches and mask boundary jitter across views. While SAM2 \cite{ravi2024sam2} incorporates a streaming memory architecture to maintain temporal coherence across continuous video frames, applying it directly to unordered multi-view collections remains ill-posed. GaussianDS resolves this bottleneck by organizing unordered camera viewpoints into a pose-aware pseudo-video sequence based on spatial proximity, establishing robust cross-view mask tracking before lifting semantics to 3D.

\subsection{Open-Vocabulary 3D Gaussian Understanding}
Extending 3DGS to open-vocabulary understanding has inspired extensive efforts to model semantic fields over 3D primitives. LangSplat \cite{langsplat_qin2024langsplat} constructs a hierarchical semantic representation by compressing multi-scale CLIP features into compact latent vectors via autoencoders. To mitigate multi-view feature discrepancies, ObjectGS \cite{zhu2025objectgs} attaches identity embeddings to Gaussians to support joint appearance reconstruction and instance segmentation. CAGS \cite{SUN2026105830} constructs local context graphs over neighboring Gaussians and enforces mask-centric contrastive learning to smooth fragmented features. Subsequent works, such as SAG3D \cite{cen2025segment} and ObjectGS \cite{zhu2025objectgs}, further scale semantic fields to instance decomposition and interactive scene editing. Crucially, most prevailing paradigms freeze pre-trained Gaussian geometry prior to semantic distillation, which precludes cross-modal feedback and leaves multi-view supervision conflicts unresolved. Departing from such decoupled schemes, GaussianDS performs end-to-end joint optimization of appearance, geometry, and semantics from scratch, explicitly anchoring 3D semantic transitions onto physical geometric depth edges.

%==========================================================================
% SECTION III: METHOD
%==========================================================================
\begin{figure*}[htp]
    \centering
    \includegraphics[width=1\linewidth]{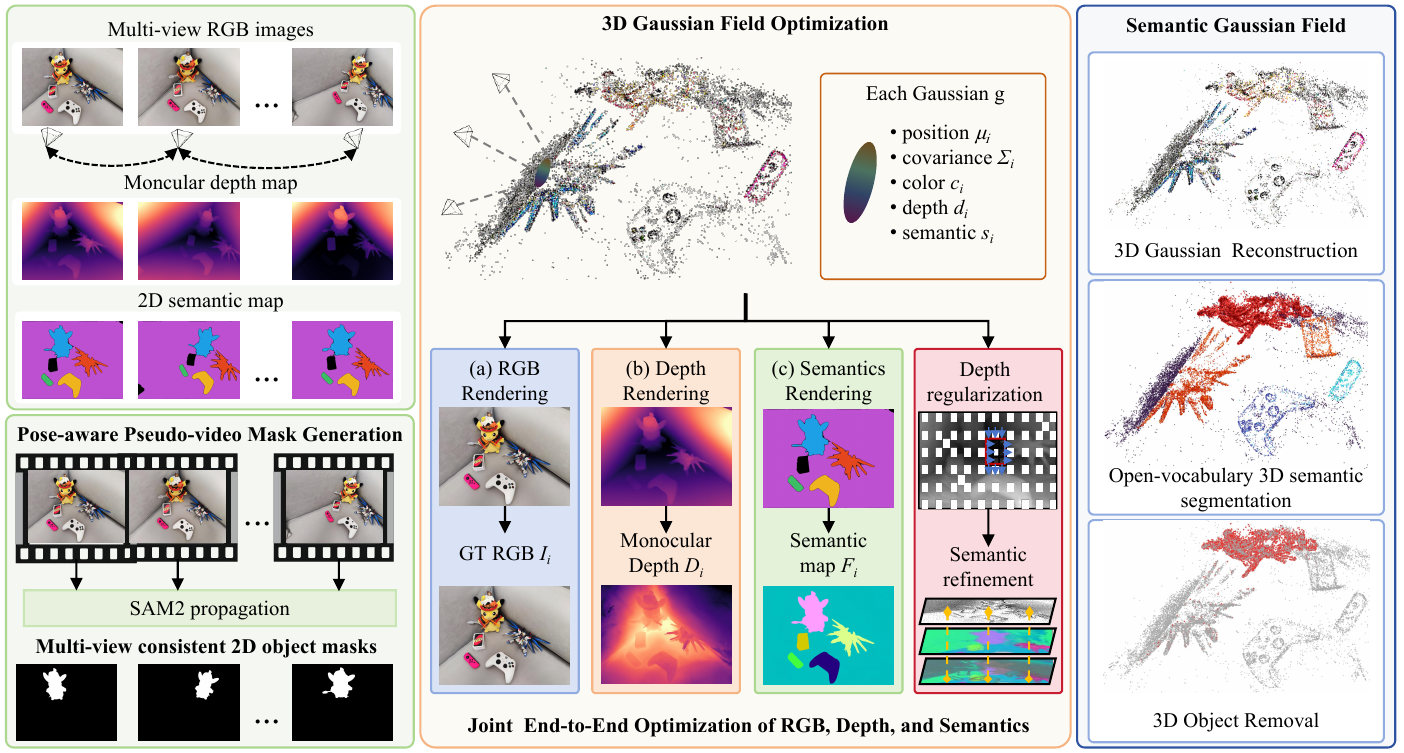}
    \caption{Overview of the GaussianDS framework. Multi-view images, monocular depth, and 2D semantic maps are converted into pose-aware pseudo-video masks, then jointly optimized through RGB, depth, semantic, and depth-edge-aware losses to obtain a semantic Gaussian field.}
    \label{fig:overview}
\end{figure*}
\section{Preliminary: 3D Gaussian Splatting}
\label{sec:preliminary}
3D Gaussian Splatting (3DGS) \cite{3dgs_kerbl20233d} parameterizes a continuous 3D scene using a collection of anisotropic 3D Gaussians $\{\mathcal{G}_i\}$. Each Gaussian is defined by its spatial center $\mu_i \in \mathbb{R}^3$, 3D covariance matrix $\Sigma_i \in \mathbb{R}^{3 \times 3}$, opacity $\alpha_i \in [0, 1]$, and spherical-harmonic appearance coefficients $c_i \in \mathbb{R}^3$. To maintain positive semi-definiteness, the covariance matrix is factorized into a rotation matrix $R_i$ and a scaling matrix $S_i$ via $\Sigma_i = R_i S_i S_i^\top R_i^\top$.

Given a camera viewpoint $\pi$, 3D Gaussians are projected onto the 2D image plane and sorted by view-space depth. Through tile-based differentiable rasterization, the rendered color $\hat{I}(u, v) \in \mathbb{R}^3$ at pixel coordinate $(u, v)$ is computed via visibility-ordered $\alpha$-compositing:
\begin{equation}
    \hat{I}(u, v) = \sum_{k=1}^K T_k \alpha_k c_k \in \mathbb{R}^3, \quad \text{with} \quad T_k = \prod_{j=1}^{k-1} (1 - \alpha_j),
    \label{eq:color_rendering}
\end{equation}
where $K$ denotes the number of depth-sorted Gaussians overlapping the ray, $\alpha_k$ is the evaluated 2D Gaussian response modulated by opacity, and $T_k$ represents the accumulated transmittance. The appearance reconstruction loss is formulated as:
\begin{equation}
    \mathcal{L}_{\mathrm{rec}} = (1 - \lambda_{\mathrm{ssim}}) \|\hat{I} - I\|_1 + \lambda_{\mathrm{ssim}} \mathcal{L}_{\mathrm{SSIM}}(\hat{I}, I).
    \label{eq:loss_rec}
\end{equation}

% ==========================================
% Section IV-A: Depth Rendering (深度属性累加)
% ==========================================
\section{Methodology}
\label{sec:method}
Our goal is to learn an open-vocabulary semantic Gaussian field from 2D foundation-model supervision without requiring dense 3D annotations. Given a collection of posed images $\mathcal{I} = \{(I_i, \pi_i)\}_{i=1}^N$ alongside estimated monocular depth maps $\{\bar{D}_i\}_{i=1}^N$, GaussianDS directly optimizes the spatial geometry, appearance, and compact latent semantics of 3D Gaussians from scratch. Unlike conventional cascaded pipelines that freeze pre-trained geometry prior to semantic distillation, our framework performs end-to-end joint optimization across rendered color $\hat{I}_i$, depth $\hat{D}_i$, and semantic feature maps $\hat{S}_i$ under each viewpoint $\pi_i$. To overcome the cross-view inconsistency and spatial boundary ambiguity inherent in 2D foundation models, GaussianDS is built upon two core designs: (1) \textit{pose-aware pseudo-video mask propagation}, which orders multi-view camera poses by spatial proximity so that SAM2 propagates view-consistent object masks across overlapping viewpoints to construct compact semantic targets, and (2) \textit{joint geometric-semantic optimization}, which simultaneously optimizes appearance, scale-shift-aligned depth, and semantic attributes, utilizing depth regularization and depth-edge constraints to anchor semantic boundaries onto physical scene geometry. Fig.~\ref{fig:overview} illustrates the overall architecture of GaussianDS.

\subsection{Depth Rendering and Geometric Regularization}
\label{sec:depth_rendering}
Vanilla 3DGS optimizes scene representations purely through apperance reconstruction objectives, lacking direct pixel-level geometric constraints. Consequently, the spatial distribution of Gaussians can deviate from actual underlying surfaces, particularly across sparsely observed or textureless regions. To impose dense geometric constraints, we incorporate camera-space $z$-depth, defined as the perpendicular distance from Gaussian centers to the camera plane under standard pinhole projection, as an auxiliary supervisory signal complementary to appearance reconstruction.

\textbf{Monocular Depth Alignment and Rendering.}
For each training view $I_i \in \mathcal{I}$, we leverage a pre-trained Depth Anything model \cite{depth_anything_yang2024depth} to obtain an estimated relative depth map $\bar{D}_i$. Following the unified ray-wise compositing formulation in Eq.~\eqref{eq:color_rendering}, the rendered camera-space depth $\hat{D}(u, v) \in \mathbb{R}$ at pixel coordinate $(u, v)$ is computed by accumulating the depth attributes of contributing Gaussians along the viewing ray:
\begin{equation}
    \hat{D}(u, v) = \sum_{k=1}^K T_k \alpha_k z_k \in \mathbb{R}, \quad \text{with} \quad T_k = \prod_{j=1}^{k-1} (1 - \alpha_j),
    \label{eq:depth_rendering}
\end{equation}
where $K$ denotes the number of sorted Gaussians contributing to the ray, $z_k \in \mathbb{R}$ is the camera-space $z$-coordinate of the $k$-th Gaussian center, $\alpha_k$ is the evaluated opacity, and $T_k$ represents the accumulated transmittance. Because monocular depth predictions are scale-ambiguous and shift-variant, directly computing Euclidean distances against rendered metric depth is ill-posed. We therefore resolve view-specific scale $s^*$ and shift $t^*$ by optimizing:
\begin{equation}
    s^*, t^* = \arg\min_{s, t} \|\hat{D} - (s \bar{D} + t)\|_1.
    \label{eq:scale_shift}
\end{equation}
The scale-shift-aligned depth loss is then formulated as:
\begin{equation}
    \mathcal{L}_{\mathrm{depth}} = \|\hat{D} - (s^* \bar{D} + t^*)\|_1.
    \label{eq:loss_depth}
\end{equation}
The optimal alignment parameters $s^*$ and $t^*$ are solved per iteration via a closed-form least-squares analytical solution across valid depth pixels without gradient backpropagation. This term aligns global scene scale while preserving the relative geometric structures predicted by foundation models.

\begin{figure}[t]
    \centering
    \includegraphics[width=0.8\linewidth]{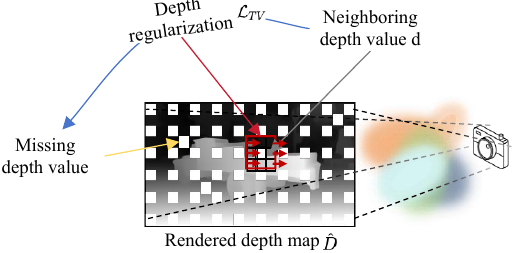}
    \caption{Effect of depth total-variation regularization. The regularizer suppresses sparse rendered-depth artifacts while preserving coarse geometric structure.}
    \label{fig-depth-reg}
\end{figure}

\textbf{Rendered Depth Total-Variation Regularization.}
While aligned monocular depth provides robust monotonic ordering priors, discrete Gaussian distributions in weakly observed regions often cause localized depth jitter, see-through artifacts, or discontinuous surface holes. To penalize high-frequency depth noise while preserving true geometric discontinuities, we enforce total variation (TV) regularization over the rendered depth map:
\begin{equation}
    \mathcal{L}_{\mathrm{tv}} = \sum_{u, v} \sqrt{\left(\hat{D}_{u+1, v} - \hat{D}_{u, v}\right)^2 + \left(\hat{D}_{u, v+1} - \hat{D}_{u, v}\right)^2},
    \label{eq:loss_tv}
\end{equation}
where $\hat{D}_{u, v}$ represents the rendered depth value at pixel coordinate $(u, v)$. As illustrated in Fig.~\ref{fig-depth-reg}, $\mathcal{L}_{\mathrm{tv}}$ enforces piecewise smoothness across locally under-constrained surfaces, acting in synergy with $\mathcal{L}_{\mathrm{depth}}$ to establish a well-disciplined geometric backbone for semantic field learning.

\subsection{Consistent Semantic Learning and Depth-Edge-Aware Refinement}
\label{sec:semantic_learning}
Lifting 2D foundation-model predictions into 3D space frequently introduces conflicting supervision signals due to viewpoint-dependent appearance shifts and spatial boundary jitter across unaligned views. Backpropagating these inconsistent targets into shared 3D Gaussian primitives leads to severe semantic drift and boundary leakage. To overcome these challenges, we introduce a pose-aware mask propagation scheme to enforce cross-view semantic coherence before lifting, coupled with a depth-edge-aware refinement objective that anchors semantic transitions onto physical geometric discontinuities during joint optimization.

\textbf{Pose-Aware Pseudo-Video Mask Propagation.}
Although SAM2 \cite{ravi2024sam2} incorporates a streaming memory mechanism to maintain temporal consistency across continuous video streams, multi-view image collections lack natural temporal continuity. To bridge this gap, we organize unordered camera viewpoints into a pose-aware pseudo-video trajectory by sorting camera positions according to Euclidean spatial proximity. This sequencing maximizes visual overlap between adjacent frames, thereby providing a physically plausible trajectory for video segmentation. Once initialized on reference frames, SAM2 propagates instance masks across the ordered trajectory, generating spatiotemporally coherent masklets that preserve consistent object identities across overlapping viewpoints.

\textbf{Compact Semantic Latent Construction.}
Given the propagated object masks, we extract high-dimensional semantic descriptors using a pre-trained OpenCLIP ViT-B/16 encoder \cite{clip_radford2021learning} applied to masked object crops. Directly optimizing high-dimensional CLIP embeddings across millions of Gaussians imposes prohibitive memory and computational footprints. To address this bottleneck, we employ a lightweight autoencoder pre-trained on object feature crops to compress the 512-dimensional CLIP embeddings into a compact $d$-dimensional latent code (we set $d=3$). These compact latent vectors serve as dense supervisory targets $S \in \mathbb{R}^{H \times W \times d}$, preserving object-level discriminative semantics while significantly reducing rendering memory overhead.

\textbf{Semantic Gaussian Rendering.}
To represent continuous semantic fields, we augment each 3D Gaussian primitive $\mathcal{G}_i$ with a learnable latent semantic vector $s_i \in \mathbb{R}^d$. Following visibility-ordered $\alpha$-compositing, the rendered semantic feature $\hat{S}(u, v) \in \mathbb{R}^d$ at coordinate $(u, v)$ is computed by accumulating the semantic attributes along the ray:
\begin{equation}
    \hat{S}(u, v) = \sum_{k=1}^K T_k \alpha_k s_k \in \mathbb{R}^d,
    \label{eq:semantic_rendering}
\end{equation}
where $s_k \in \mathbb{R}^d$ denotes the semantic attribute of the $k$-th Gaussian. The base semantic rendering loss is formulated as:
\begin{equation}
    \mathcal{L}_{\mathrm{sem}} = \|\hat{S} - S\|_1,
    \label{eq:loss_sem}
\end{equation}
where $S$ denotes the target latent semantic map obtained from the masked CLIP features.

\textbf{Depth-Edge-Aware Semantic Refinement.}
Despite cross-view mask propagation, lifted 2D semantic supervision inherently exhibits ambiguous contours in complex occlusion regions, which causes semantic features to bleed across physical object margins. Because semantic transitions in natural scenes strongly correlate with geometric surface boundaries, we introduce a depth-edge-aware refinement loss that explicitly anchors semantic boundaries to physical geometric discontinuities. Specifically, we apply a Canny edge detector to the scale-shift-aligned monocular depth map to construct a binary depth-edge mask $E_d = \mathrm{Canny}(s^* \bar{D} + t^*)$, where $E_d(u,v) = 1$ indicates a physical geometric discontinuity. We then formulate the boundary refinement loss by spatially weighting the semantic discrepancy:
\begin{equation}
    \mathcal{L}_{\mathrm{ds}} = \|E_d \odot (\hat{S} - S)\|_1,
    \label{eq:loss_ds}
\end{equation}
where $\odot$ denotes the Hadamard product. By penalizing semantic prediction errors specifically along depth discontinuities, $\mathcal{L}_{\mathrm{ds}}$ guides the joint optimization to align 3D semantic boundaries with underlying scene geometry, effectively eliminating semantic bleeding and sharpening object extents.

\subsection{Joint Optimization Objective}
\label{sec:loss}
Rather than adopting a multi-stage training routine that isolates semantic distillation from geometric reconstruction, GaussianDS unifies all supervision signals into a single-stage joint optimization framework from scratch. The overall optimization objective is defined as a weighted combination of appearance, depth, and semantic losses:
\begin{equation}
    \mathcal{L}_{\mathrm{render}} = \mathcal{L}_{\mathrm{rec}} + \lambda_{\mathrm{depth}} \mathcal{L}_{\mathrm{depth}} + \lambda_{\mathrm{tv}} \mathcal{L}_{\mathrm{tv}} + \lambda_{\mathrm{sem}} \mathcal{L}_{\mathrm{sem}} + \lambda_{\mathrm{ds}} \mathcal{L}_{\mathrm{ds}},
    \label{eq:total_loss}
\end{equation}
where $\lambda_{\mathrm{depth}}$, $\lambda_{\mathrm{tv}}$, $\lambda_{\mathrm{sem}}$, and $\lambda_{\mathrm{ds}}$ are balancing hyperparameters governing geometric regularization and semantic alignment.

\subsection{Open-Vocabulary Query and Downstream Editing}
\label{sec:semantic_query}
\textbf{Open-Vocabulary Query and Segmentation.}
Given an arbitrary text query, we encode the textual prompt using the CLIP text encoder and map it into the $d$-dimensional latent semantic space via the pre-trained autoencoder encoder, obtaining the query embedding $\phi_Q \in \mathbb{R}^d$. The semantic similarity between the query embedding and a Gaussian semantic attribute $\phi_G = s_i$ (or a rendered pixel feature $\hat{S}(u,v)$) is evaluated via cosine similarity:
\begin{equation}
    \cos(\phi_Q, \phi_G) = \frac{\phi_Q \cdot \phi_G}{\|\phi_Q\|_2 \|\phi_G\|_2}.
    \label{eq:cosine_sim}
\end{equation}
To perform 2D novel-view semantic segmentation, we render the dense semantic feature map $\hat{S}$, compute the pixel-wise cosine similarity against $\phi_Q$, and apply a threshold to generate the binary segmentation mask.

\begin{figure*}[ht]
    \centering
    \includegraphics[width=1\linewidth]{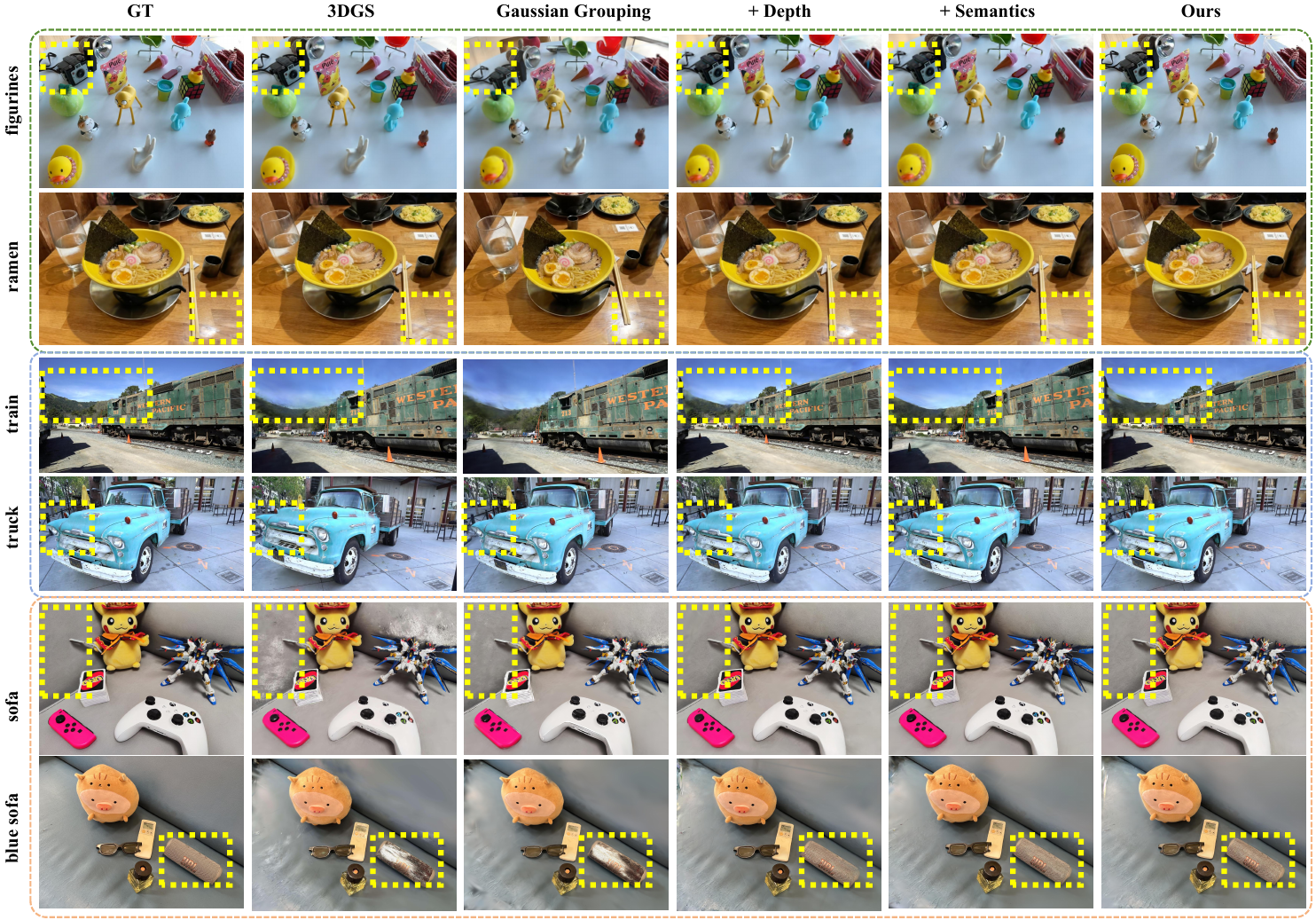}
    \caption{Qualitative comparisons of 3D reconstruction on the LERF\cite{lerf2023}, TanDB\cite{3dgs_kerbl20233d} and 3D-OVS datasets~\cite{liu2023weakly3d_ovs}.}
    \label{fig-rec-v1}
\end{figure*}
\textbf{3D Object Removal and Scene Decomposition.}
To execute 3D object removal, we assign Gaussian primitives to semantic categories by evaluating cosine similarity against target query embeddings or class prototypes. Gaussians whose semantic similarity exceeds a predetermined threshold are classified as belonging to the target object and subsequently pruned from the Gaussian collection. Because each Gaussian retains its intrinsic geometric and appearance attributes independently, object removal can be executed directly by pruning target primitives without modifying the appearance or structural properties of the remaining scene.

%==========================================================================
% SECTION IV: EXPERIMENTS
%==========================================================================
\section{Experiments}
\label{sec:experiments}

\subsection{Experimental Setup}
\label{sec:exp_setup}
\textbf{Datasets and Evaluation Metrics.}
We evaluate GaussianDS across four benchmark datasets: LERF \cite{lerf2023}, 3D-OVS \cite{liu2023weakly3d_ovs}, and standard 3DGS scenes from Tanks and Temples and Deep Blending \cite{3dgs_kerbl20233d}. Novel-view 3D reconstruction quality is assessed via PSNR, SSIM, and LPIPS on held-out test viewpoints. For open-vocabulary 3D semantic segmentation, text queries are mapped into the latent semantic space, and binary foreground masks are obtained by thresholding rendered cosine similarity maps to compute mean Intersection-over-Union (mIoU) against annotated ground truth. To rigorously validate boundary localization fidelity on 3D-OVS, we additionally evaluate mean Boundary IoU (mBIoU) restricted within a narrow morphological dilation band around object contours. Downstream 3D object removal is further evaluated on 3D-OVS editing scenarios.

\textbf{Implementation Details.}
We adopt OpenCLIP ViT-B/16 \cite{clip_radford2021learning} to extract 512-dimensional semantic embeddings from masked object crops resized to $224 \times 224$ pixels. Cross-view mask propagation across pose-ordered pseudo-video sequences is conducted using SAM2 with the \texttt{sam2-hiera-large} model \cite{ravi2024sam2}. An offline-trained, frozen lightweight autoencoder compresses the 512-dimensional CLIP descriptors into a $d=3$ dimensional latent code attached to Gaussian primitives. Customized CUDA rasterizers are implemented for differentiable depth and semantic rendering. The loss balancing weights are set to $\lambda_{\mathrm{depth}} = 0.01$, $\lambda_{\mathrm{tv}} = 1.5$, $\lambda_{\mathrm{sem}} = 5.0$, and $\lambda_{\mathrm{ds}} = 1.0$. All parameters across appearance, depth, and semantics are optimized jointly from scratch for 30K iterations on a single NVIDIA RTX 4090 GPU.

%--------------------------------------------------------------------------
\subsection{Multi-view 3D Reconstruction}
%--------------------------------------------------------------------------
We evaluate multi-view 3D reconstruction across three benchmark suites: 3D-OVS~\cite{liu2023weakly3d_ovs}, Tanks and Temples / Deep Blending (TanDB)~\cite{3dgs_kerbl20233d}, and LERF~\cite{lerf2023}. As reported in Table~\ref{tab-rec-main}, GaussianDS consistently outperforms both vanilla 3DGS and Gaussian Grouping across the majority of evaluated scenes. This advantage is particularly pronounced on 3D-OVS, where limited viewpoints render unconstrained RGB reconstruction ill-posed. By coupling scale-shift-aligned monocular depth priors with total-variation regularization, GaussianDS provides crucial geometric constraints that stabilize Gaussian spatial distributions in under-constrained regions. As demonstrated in Fig.~\ref{fig-rec-v1}, our method faithfully recovers fine-grained structural contours across specular surfaces and geometrically ambiguous boundaries. Furthermore, parameterizing semantics with compact $d=3$ latent codes introduces negligible per-primitive memory overhead, allowing GaussianDS to maintain an average real-time rendering speed comparable to vanilla 3DGS. In contrast, assigning high-dimensional identity embeddings, as in Gaussian Grouping, compromises both reconstruction quality and rendering frame rates.

\begin{figure}[ht]
    \centering
    \includegraphics[width=0.8\linewidth]{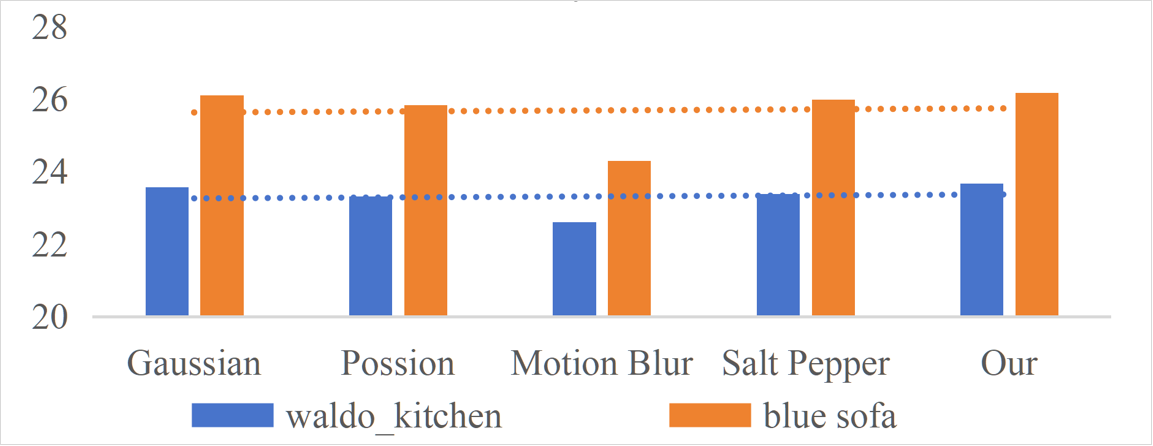}
    \caption{Reconstruction results under different types of depth noise on \textit{waldo\_kitchen} and \textit{blue sofa}.}
    \label{fig-noise}
\end{figure}

\begin{table*}[htb]
    \centering
    \caption{Reconstruction results on 3D-OVS, Tanks and Temples/Deep Blending, and LERF. Each scene reports SSIM, PSNR, LPIPS, and FPS.}
    \resizebox{1\linewidth}{!}{
    \begin{tabular}{l|cccc|cccc|cccc}
    \toprule
    Scene & \multicolumn{4}{c|}{\textbf{3DGS}} & \multicolumn{4}{c|}{\textbf{Gaussian Grouping}} & \multicolumn{4}{c}{\textbf{GaussianDS}} \\
    Metric & SSIM$\uparrow$ & PSNR$\uparrow$ & LPIPS$\downarrow$ & FPS & SSIM$\uparrow$ & PSNR$\uparrow$ & LPIPS$\downarrow$ & FPS & SSIM$\uparrow$ & PSNR$\uparrow$ & LPIPS$\downarrow$ & FPS \\
    \midrule
    \multicolumn{13}{l}{\textbf{3D-OVS}} \\
    bench           & 0.42  & 16.02  & 0.50  & $\sim$195 & 0.53 & 18.71 & 0.42 & $\sim$ 170     & \textbf{0.72}  & \textbf{24.81} & \textbf{0.15} & $\sim$190 \\
    blue sofa       & 0.60  & 20.65  & 0.58  & --        & 0.57 &20.92 & 0.58  & --     & \textbf{0.76}  & \textbf{26.18} & \textbf{0.44} & -- \\
    rooms           & 0.80  & 23.80  & 0.38  & --        & 0.79  &23.75  & 0.41  & --     & \textbf{0.83}  & \textbf{28.48} & \textbf{0.23} & -- \\
snacks          & 0.78  & 23.48  & 0.47  & --        & 0.77 & 23.17 & 0.47 & --        & \textbf{0.88}  & \textbf{30.07} & \textbf{0.26} & -- \\
office\_desk    & 0.64  & 17.48  & 0.41  & --        & 0.71 & 22.19 & 0.32 & --        & \textbf{0.81}  & \textbf{24.74} & \textbf{0.22} & -- \\
lawn            & 0.29  & 10.98  & 0.52  & --        & 0.26 & 10.69 & 0.53 & --        & \textbf{0.69}  & \textbf{18.54} & \textbf{0.20} & -- \\
table           & 0.84  & 20.12  & 0.33  & --        & 0.79 & 19.84 & 0.54 & --        & \textbf{0.91}  & \textbf{25.88} & \textbf{0.25} & -- \\
sofa            & 0.59  & 20.75  & 0.52  & --        & 0.59 & 20.46 & 0.55 & --        & \textbf{0.80}  & \textbf{26.13} & \textbf{0.29} & -- \\
bed             & 0.72  & 21.35  & 0.44  & --        & 0.71 & 21.37 & 0.46 & --        & \textbf{0.85}  & \textbf{24.67} & \textbf{0.14} & -- \\
covered\_desk   & 0.76  & 21.94  & 0.45  & --        & 0.82 & 24.62 & 0.39 & --        & \textbf{0.86}  & \textbf{27.33} & \textbf{0.31} & -- \\
\midrule
\multicolumn{13}{l}{\textbf{TanDB}} \\
train           & 0.80 & 21.09 & 0.21 & $\sim$154 & 0.81 & 21.81 & 0.22 & $\sim$126 & \textbf{0.87} & \textbf{22.04} & \textbf{0.18} & $\sim$145 \\
truck           & 0.88 & 25.19 & 0.15 & --        & 0.88 & 24.92 & 0.20 & --        & \textbf{0.89} & \textbf{25.54} & \textbf{0.15} & -- \\
Dr Johnson      & 0.90 & 28.77 & 0.24 & --        & 0.90 & 28.46 & 0.29 & --        & \textbf{0.92} & \textbf{30.51} & \textbf{0.21} & -- \\
Playroom        & 0.91 & 30.04 & 0.24 & --        & 0.93 & 29.75 & 0.22 & --        & \textbf{0.94} & \textbf{30.48} & \textbf{0.19} & -- \\
    \midrule
    \multicolumn{13}{l}{\textbf{LERF}} \\
    figurines       & 0.82  & 22.81  & 0.24  & $\sim$277 &0.82  & 24.96   & \textbf{0.17}     &$\sim$260     & \textbf{0.85}  & \textbf{25.00} & 0.20 & $\sim$257 \\
    ramen           & 0.86  & 24.90  & \textbf{0.17}  & --   & 0.83 & 24.11 & 0.30      & --     & \textbf{0.87}  & \textbf{25.10} & 0.19 & -- \\
    teatime         & 0.82  & 22.90  & 0.29  & --  & 0.91   & 30.73  & 0.18 & --     & \textbf{0.93}  & \textbf{33.96} & \textbf{0.17} & -- \\
    waldo\_kitchen  & \textbf{0.87}  & 21.97  & 0.22  & --  & 0.86 &  \textbf{32.84}   & 0.23   & --     & \textbf{0.87}           & 23.67 & \textbf{0.20} & -- \\
    \bottomrule
    \end{tabular}
    }
    \label{tab-rec-main}
\end{table*}
\textbf{Robustness to Depth Noise Perturbations.}
To evaluate system resilience against imperfect monocular depth estimators, we inject synthetic perturbations into the input depth maps, including Gaussian noise, Poisson noise, motion blur, and salt-and-pepper noise. As shown in Fig.~\ref{fig-noise}, reconstruction metrics remain remarkably robust across standard noise distributions. Motion blur causes the most noticeable degradation because it smooths localized depth gradients and attenuates geometric edges detected by the Canny operator, thereby degrading the precision of the boundary refinement loss $\mathcal{L}_{\mathrm{ds}}$. This empirical finding validates our core design principle: GaussianDS relies primarily on relative structural depth discontinuities rather than absolute metric depth accuracy.

%--------------------------------------------------------------------------
\subsection{Open-Vocabulary 3D Semantic Segmentation}
\label{sec:semantic_segmentation}
We evaluate open-vocabulary 3D semantic segmentation on the LERF benchmark \cite{lerf2023} to examine whether rendered Gaussian semantic fields support precise zero-shot text-queried localization. We compare GaussianDS against leading semantic radiance and Gaussian frameworks, including LERF \cite{lerf2023}, Gaussian Grouping \cite{gg_gaussian_grouping}, LangSplat \cite{langsplat_qin2024langsplat}, LangSplatv2 \cite{li2025langsplatv2}, OpenGaussian \cite{opengaussian}, SparseLGS \cite{hu2024sparselgs}, GAGS \cite{peng2026gags}, and GOI \cite{qu2024goi}.

\begin{table}[ht]
    \centering
    \caption{Quantitative comparison of open-vocabulary semantic segmentation (mIoU, \%) on the LERF dataset \cite{lerf2023}.}
    \begin{tabular}{llllll}
    \toprule
        \textbf{Method}     & \textbf{figurines} & \textbf{ramen} & \textbf{teatime} & \textbf{kitchen} & \textbf{overall} \\ \midrule
        LERF          & 33.5               & 28.3           & 49.7             & 37.9             & 37.4             \\
        GS-Grouping & 34.6  & 26.4 & 32.3 & 34.6 & 36.6 \\
        LangSplat     & 44.7               & 51.2           & 65.1             & 44.5             & 51.4             \\
        LangSplatv2    & 56.4               & 51.8           & \textbf{72.2}    & 59.1             & 59.9            \\
        OpenGaussian & 39.3              & 31.0          & 60.4            & 22.7            & 38.4            \\
        SparseLGS    & 47.3              & 55.2          & 40.7            & 22.9            & 41.5            \\
        GAGS    & 53.6              & 46.8          & 60.3            & 55.8            & 54.1   \\
        GOI    & 44.7               & 51.2           & 65.1             & 44.5             & 51.4 \\
        
        \textbf{GaussianDS} & \textbf{57.0}     & \textbf{56.7} & 67.0             & \textbf{61.4}   & \textbf{60.5}   \\ \bottomrule
    \end{tabular}
\label{tab-seg}
\end{table}

\begin{figure}[ht]
    \centering
    \includegraphics[width=1\linewidth]{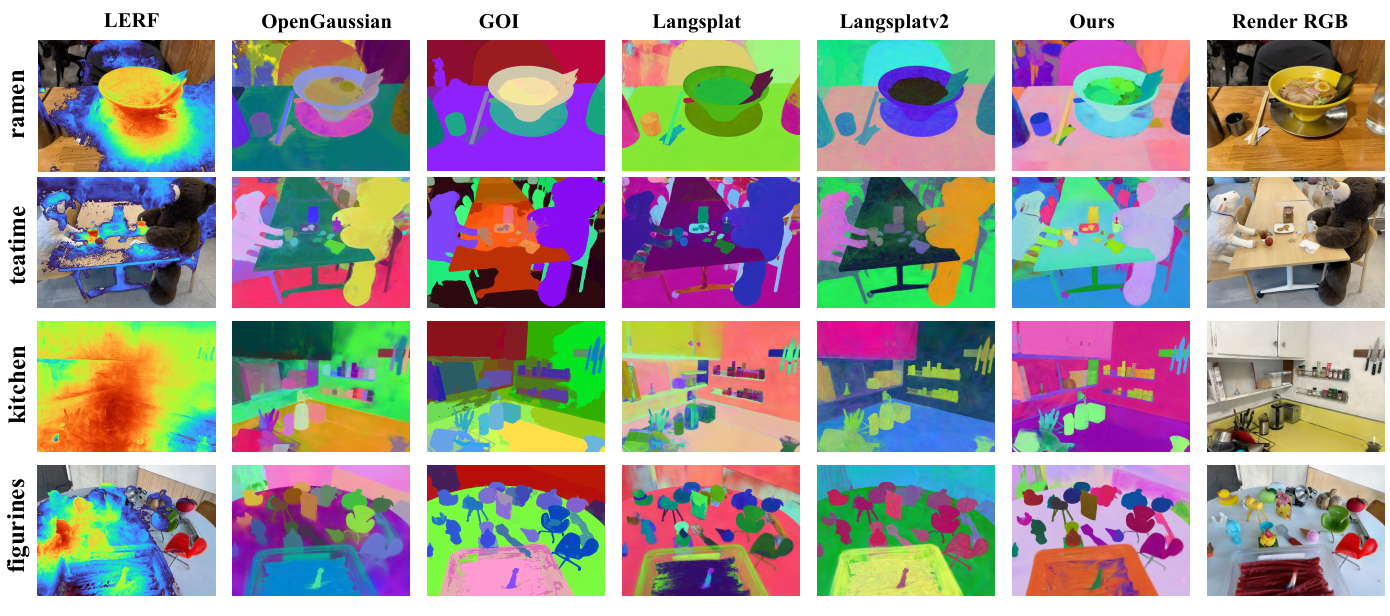}
    \caption{Qualitative comparisons of 3D semantic segmentation on the LERF dataset \cite{lerf2023}.}
    \label{fig-seg-lerf}
\end{figure}

As reported in Table~\ref{tab-seg}, GaussianDS establishes the highest overall segmentation accuracy among all evaluated baselines, achieving $60.5\%$ overall mIoU and outperforming the strongest baseline LangSplatv2 ($59.9\%$).  Notably, GaussianDS achieves substantial gains in heavily cluttered environments, improving mIoU to $61.4\%$ on \textit{waldo kitchen} and $56.7\%$ on \textit{ramen}. As qualitatively visualized in Fig.~\ref{fig-seg-lerf}, while baseline methods suffer from cross-view supervision conflicts that cause semantic features to bleed into adjacent background structures, our geometry-aligned supervision strictly bounds semantic transitions to physical object contours.

\subsection{Boundary-Aware Semantic Segmentation}
\label{sec:boundary_segmentation}
While standard mean Intersection-over-Union (mIoU) measures overall regional mask overlap, it is predominantly governed by interior object pixels and remains less sensitive to subtle semantic bleeding along spatial margins. To explicitly examine whether the rendered Gaussian semantics adhere to physical object contours rather than merely classifying bulk interior regions, we conduct boundary-sensitive evaluation on the 3D-OVS dataset \cite{liu2023weakly3d_ovs}, reporting both region-level mIoU and boundary-level mBIoU.

\begin{table*}[ht]
\centering
    \caption{Quantitative results of boundary-aware semantic segmentation on 3D-OVS.}
\resizebox{\linewidth}{!}{
\begin{tabular}{l|cc cc cc cc cc cc}
\toprule
Method & \multicolumn{2}{c}{bed} & \multicolumn{2}{c}{bench} & \multicolumn{2}{c}{room} & \multicolumn{2}{c}{sofa} & \multicolumn{2}{c}{lawn} & \multicolumn{2}{c}{Overall} \\ \midrule
Metric & mIoU & mBIoU & mIoU & mBIoU & mIoU & mBIoU & mIoU & mBIoU & mIoU & mBIoU & mIoU & mBIoU \\
\midrule
GAGS & 91.39 & 81.52 & 94.29 & 88.48 & 86.58 & 80.89 & 86.35 & 80.25 & 91.25 & 84.22 & 89.97 & 83.07 \\
LangSplatV2 & 93.01 & 82.53 & 94.93 & 89.59 & 96.15 & 90.56 & 92.33 & 84.73 & 96.58  & 93.60 & 94.60 & 88.40 \\
OpenGaussian & 89.65 & 81.10 & 90.14 & 76.13 & 82.11 & 73.15 & 88.48 & 83.60 & 84.02 & 79.25 & 86.88 & 78.65 \\
ReferSplat & 88.36 & 82.67 & 92.74 & 83.45 & 90.59 & 82.08 & 95.09 & 81.28 & 95.95 & 84.04 & 92.55 & 84.70 \\
\textbf{GaussianDS} & \textbf{97.86} & \textbf{86.38} & \textbf{96.97} & \textbf{91.99} & \textbf{99.05} & \textbf{93.38} & \textbf{98.06} & \textbf{85.75} & \textbf{97.00} & \textbf{93.91} & \textbf{97.79} & \textbf{90.28} \\
\bottomrule
\end{tabular}
   }
\label{tab-sem-3dovs}
\end{table*}

\textbf{Quantitative Boundary Evaluation.}
As summarized in Table~\ref{tab-sem-3dovs}, GaussianDS establishes state-of-the-art accuracy across all five evaluated 3D-OVS scenes, achieving $97.79\%$ overall mIoU and $90.28\%$ overall mBIoU, outperforming the leading baseline LangSplatV2 by $3.19\%$ and $1.88\%$ respectively. The region-level mIoU improvements confirm superior overall semantic assignment, notably on complex scenes such as \textit{room} ($99.05\%$) and \textit{bed} ($97.86\%$). More crucially, GaussianDS secures the highest mBIoU across every individual scene, validating that depth-edge-aware refinement effectively penalizes semantic bleeding and sharpens object margins along physical boundaries.

\begin{figure}[ht]
    \centering
    \includegraphics[width=1\linewidth]{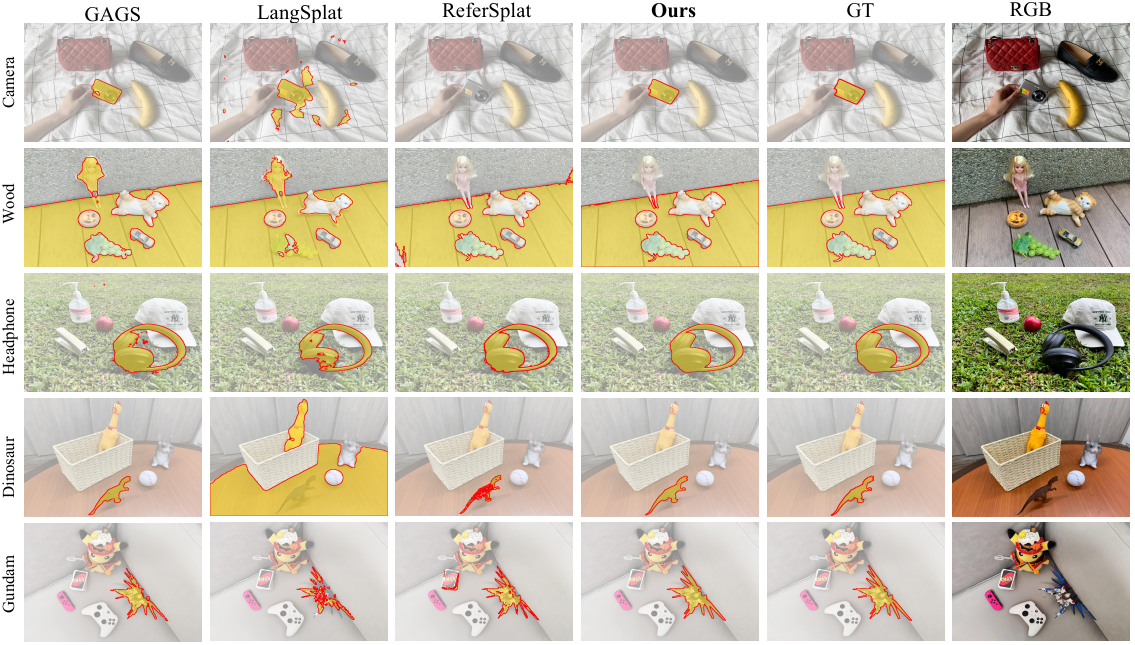}
    \caption{Qualitative comparisons of open-vocabulary 3D semantic segmentation on the 3D-OVS dataset.}
    \label{fig-boundary-3dovs}
\end{figure}

\textbf{Qualitative Boundary Analysis.}
Figure~\ref{fig-boundary-3dovs} provides qualitative boundary-focused comparisons across representative objects. Competing baselines frequently suffer from fragmented object masks or excessive semantic dilation that erroneously merges foreground semantics with neighboring background structures. In contrast, GaussianDS reconstructs sharp, topologically coherent semantic boundaries that faithfully align with physical geometric extents, providing direct visual evidence for the efficacy of our depth-edge constraint.

\subsection{Ablation Studies}
\label{sec:core-ablations}
To validate the individual contributions of our algorithmic designs, we systematically conduct ablation experiments across five key dimensions: (1) pose-aware pseudo-video mask propagation, (2) scale-shift monocular depth alignment, (3) loss component decomposition and depth-edge refinement, (4) latent semantic capacity, and (5) hyperparameter sensitivity dynamics.

\textbf{Pose-Aware Pseudo-Video Mask Propagation}
The cross-view tracking performance of SAM2 relies on visual continuity across consecutive inputs. To validate our camera proximity sequencing strategy, Table~\ref{tab-ordering} evaluates three frame-ordering schemes across the 3D-OVS scenes. We report the mean Euclidean camera-center distance between adjacent frames alongside the adjacent-frame mask IoU computed from propagated instance masks. Compared with random sequence ordering, pose-nearest ordering reduces inter-frame camera distance from $5.00$ to $1.50$ and substantially elevates adjacent mask consistency from $33.92\%$ to $53.21\%$. This confirms that pose-conditioned trajectory construction provides a stable spatiotemporal substrate that minimizes mask jitter prior to 3D lifting.

\begin{table}[ht]
\centering
\caption{Quantitative comparison of pseudo-video ordering strategies on five 3D-OVS scenes.}
\label{tab-ordering}
\resizebox{1\linewidth}{!}{
\begin{tabular}{lcc}
\toprule
\textbf{Ordering Strategy} & \textbf{Adjacent Pose Distance}$\downarrow$ & \textbf{Adjacent Mask IoU (\%)$\uparrow$} \\
\midrule
Random & 5.00 & 33.92 \\
Filename order & 2.03 & 50.85 \\
\textbf{Pose-nearest (Ours)} & \textbf{1.50} & \textbf{53.21} \\
\bottomrule
\end{tabular}
}
\end{table}

\textbf{Scale-Shift Monocular Depth Alignment}
Because monocular depth predictions exhibit inherent scale and shift ambiguities, directly minimizing uncalibrated Euclidean depth errors causes severe optimization instability. Table~\ref{tab-depth-align} compares our scale-shift formulation against raw $L_1$ supervision and min-max normalization on the 3D-OVS \textit{bench} scene under identical initializations. Direct raw $L_1$ depth supervision achieves only $21.31$ dB PSNR and $59.69\%$ mBIoU due to scale discrepancies. While min-max normalization stabilizes optimization, our scale-shift alignment in Eq.~\eqref{eq:scale_shift} achieves the highest geometric fidelity and semantic consistency, reaching $24.81$ dB PSNR and $91.99\%$ mBIoU, justifying the necessity of the aligned depth loss $\mathcal{L}_{\mathrm{depth}}$.

\begin{table}[ht]
\centering
\caption{Ablation study on depth supervision formulations on \textit{bench}.}
\label{tab-depth-align}
\resizebox{\linewidth}{!}{
\begin{tabular}{lccccc}
\toprule
\textbf{Depth Supervision Formulation} & \textbf{SSIM}$\uparrow$ & \textbf{PSNR}$\uparrow$ & \textbf{LPIPS}$\downarrow$ & \textbf{mIoU (\%)$\uparrow$} & \textbf{mBIoU (\%)$\uparrow$} \\
\midrule
Raw $L_1$ supervision & 0.684 & 21.31 & 0.246 & 71.71 & 59.69 \\
Min-max normalized depth & 0.699 & 23.93 & \textbf{0.184} & 95.00 & 89.21 \\
\textbf{Scale-shift aligned depth} & \textbf{0.701} & \textbf{24.81} & \textbf{0.184} & \textbf{96.97} & \textbf{91.99} \\
\bottomrule
\end{tabular}
}
\end{table}

\textbf{Loss Component Decomposition and Boundary Refinement}
We evaluate the individual and synergistic contributions of the loss objectives defined in the joint optimization objective in Eq.~\eqref{eq:total_loss}, specifically the scale-shift depth loss $\mathcal{L}_{\mathrm{depth}}$, the rendered depth TV regularizer $\mathcal{L}_{\mathrm{tv}}$, and the depth-edge semantic refinement loss $\mathcal{L}_{\mathrm{ds}}$ in Eq.~\eqref{eq:loss_ds}. Table~\ref{tab-ablation} reports quantitative ablations on \textit{bench} and \textit{waldo kitchen}. Incorporating $\mathcal{L}_{\mathrm{depth}}$ provides the primary boost to reconstruction fidelity ($+6.44$ dB PSNR on \textit{bench}), whereas activating the depth-edge loss $\mathcal{L}_{\mathrm{ds}}$ yields decisive improvements in semantic mIoU by penalizing boundary leakage ($+37.30\%$ mIoU on \textit{bench}). Combining all three objectives achieves the highest joint reconstruction and segmentation performance, demonstrating that geometric regularization and boundary-aware semantic refinement function as complementary constraints during joint optimization.

\begin{table}[ht]
\centering
\caption{Ablation study on loss components $\mathcal{L}_{\mathrm{depth}}$, $\mathcal{L}_{\mathrm{tv}}$, and $\mathcal{L}_{\mathrm{ds}}$ on \textit{bench} and \textit{waldo kitchen}.}
\label{tab-ablation}
%\resizebox{\linewidth}{!}{
\begin{tabular}{ccc|cc|cc}
\toprule
\multicolumn{3}{c|}{\textbf{Loss Components (Eq.~\eqref{eq:total_loss})}} & \multicolumn{2}{c|}{\textit{bench}} & \multicolumn{2}{c}{\textit{waldo kitchen}} \\
$\mathcal{L}_{\mathrm{depth}}$ & $\mathcal{L}_{\mathrm{tv}}$ & $\mathcal{L}_{\mathrm{ds}}$ & PSNR$\uparrow$ & mIoU$\uparrow$ & PSNR$\uparrow$ & mIoU$\uparrow$ \\
\midrule
 & & & 16.02 & 47.10 & 21.97 & 32.33 \\
\checkmark & & & 22.46 & 78.40 & 22.55 & 47.80 \\
 & \checkmark & & 21.94 & 71.93 & 23.21 & 47.27 \\
 & & \checkmark & 18.42 & 84.40 & 21.44 & 41.51 \\
\checkmark & \checkmark & & 24.10 & 91.34 & 23.55 & 51.72 \\
\checkmark & & \checkmark & 23.73 & 93.10 & 22.75 & 59.19 \\
\midrule
\checkmark & \checkmark & \checkmark & \textbf{24.81} & \textbf{96.97} & \textbf{23.67} & \textbf{61.43} \\
\bottomrule
\end{tabular}
%}
\end{table}

\textbf{Latent Semantic Dimensionality.} Table~\ref{tab-dims} reports the segmentation accuracy and resource consumption across varying latent dimensions $d\in[3,32]$ on \textit{waldo kitchen}. Scaling $d$ from 3 to 16 progressively improves mIoU from 61.42\% to 62.83\%, whereas $d=32$ exhibits redundant capacity and slight optimization saturation (61.32\% mIoU). Moreover, higher dimensions drastically inflate memory footprints and training latency ($>210$ minutes at $d=32$). Our compact $d=3$ configuration retains 97.7\% of the peak performance while requiring only 39 minutes and 4.2 GB VRAM, making it our default setting for scalable deployment.

\begin{figure}[t]
\centering
\includegraphics[width=1\linewidth]{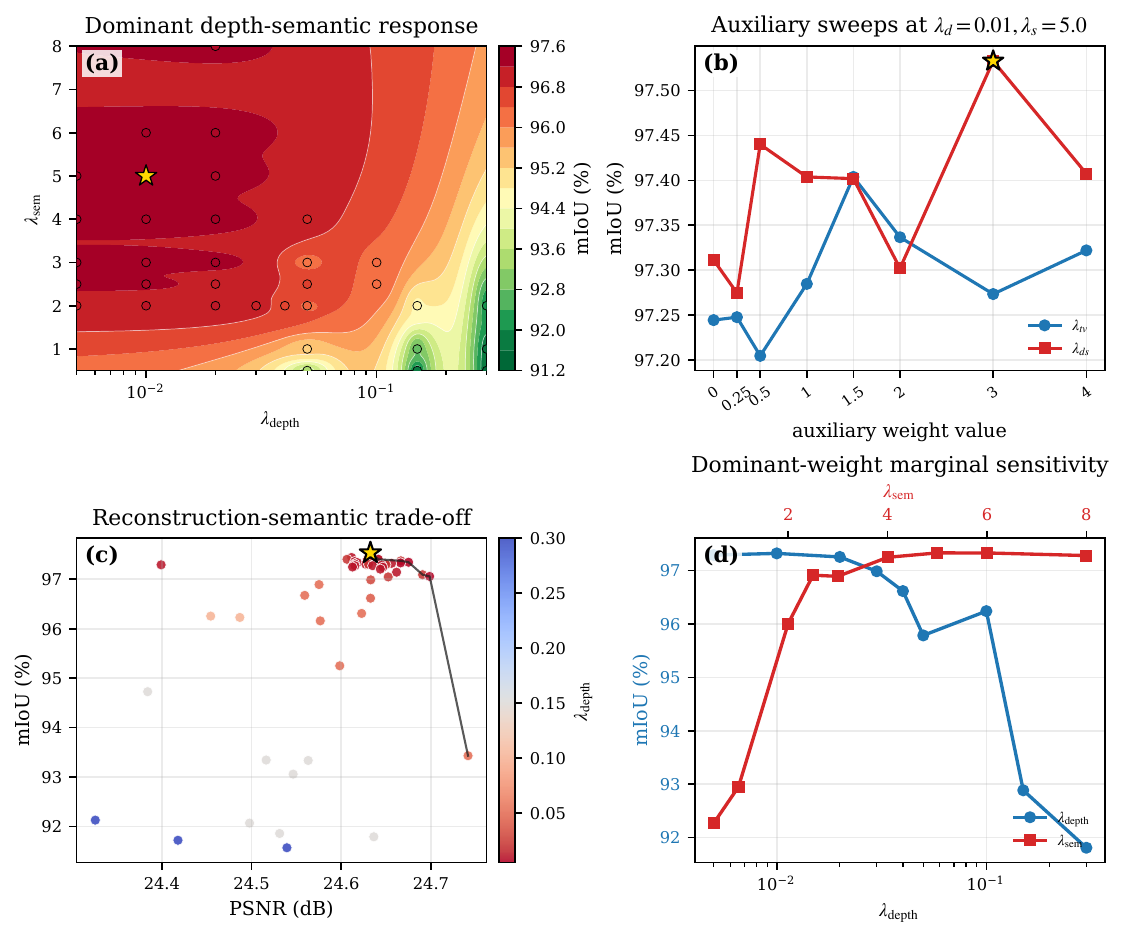}
\caption{Hyperparameter sensitivity analysis. Response surfaces highlight the trade-off between geometric depth and semantic objectives.}
\label{fig-lambda-sweep}
\end{figure}

\begin{table}[ht]
\centering
\caption{Segmentation accuracy and computational overhead across latent semantic dimensions $d$ on LERF \textit{waldo kitchen}.}
\label{tab-dims}
\resizebox{\linewidth}{!}{
\begin{tabular}{l|cccccc|c}
\toprule
\textbf{Latent Dimension ($d$)} & \textbf{3} & \textbf{6} & \textbf{8} & \textbf{12} & \textbf{16} & \textbf{32} & \textbf{3DGS} \\
\midrule
Training Time (min) & $>39$ & $>51$ & $>66$ & $>92$ & $>123$ & $>210$ & $> \textbf{20}$ \\
VRAM (GB) & $\sim 4.2$ & $\sim 5.0$ & $\sim 7.9$ & $\sim 8.3$ & $\sim 8.5$ & $\sim 10.0$ & $\sim \textbf{3.0}$ \\
mIoU (\%) & 61.42 & 61.75 & 62.13 & 62.81 & \textbf{62.83} & 61.32 & $\text{--}$ \\
PSNR (dB) & 23.67 & 23.68 & 23.67 & \textbf{23.70} & 23.69 & 23.64 & 21.97 \\
\bottomrule
\end{tabular}
}
\end{table}

\textbf{Sensitivity analysis on Multi-Task Loss Weights.}
We conduct an extensive sensitivity sweep over the multi-task loss balancing weights in Eq.~\eqref{eq:total_loss}, visualized in Fig.~\ref{fig-lambda-sweep}. The depth weight $\lambda_{\mathrm{depth}}$ exhibits a strong negative correlation with semantic mIoU ($r = -0.89$), where setting $\lambda_{\mathrm{depth}} > 0.1$ degrades mIoU toward $0.92$, indicating that excessive geometric gradients suppress latent semantic adaptation. Conversely, semantic weight $\lambda_{\mathrm{sem}}$ correlates positively with mIoU ($r = +0.69$), saturating within $[4.0, 6.0]$. Auxiliary weights $\lambda_{\mathrm{tv}}$ and $\lambda_{\mathrm{ds}}$ demonstrate broad stability basins around $1.5$ and $1.0$. Consequently, configuring $\lambda_{\mathrm{depth}} \in [0.01, 0.02]$ and $\lambda_{\mathrm{sem}} \in [4.0, 5.0]$ achieves the optimal trade-off between geometric stability and semantic sharpness.

\begin{figure}[!ht]
    \centering
    \includegraphics[width=1\linewidth]{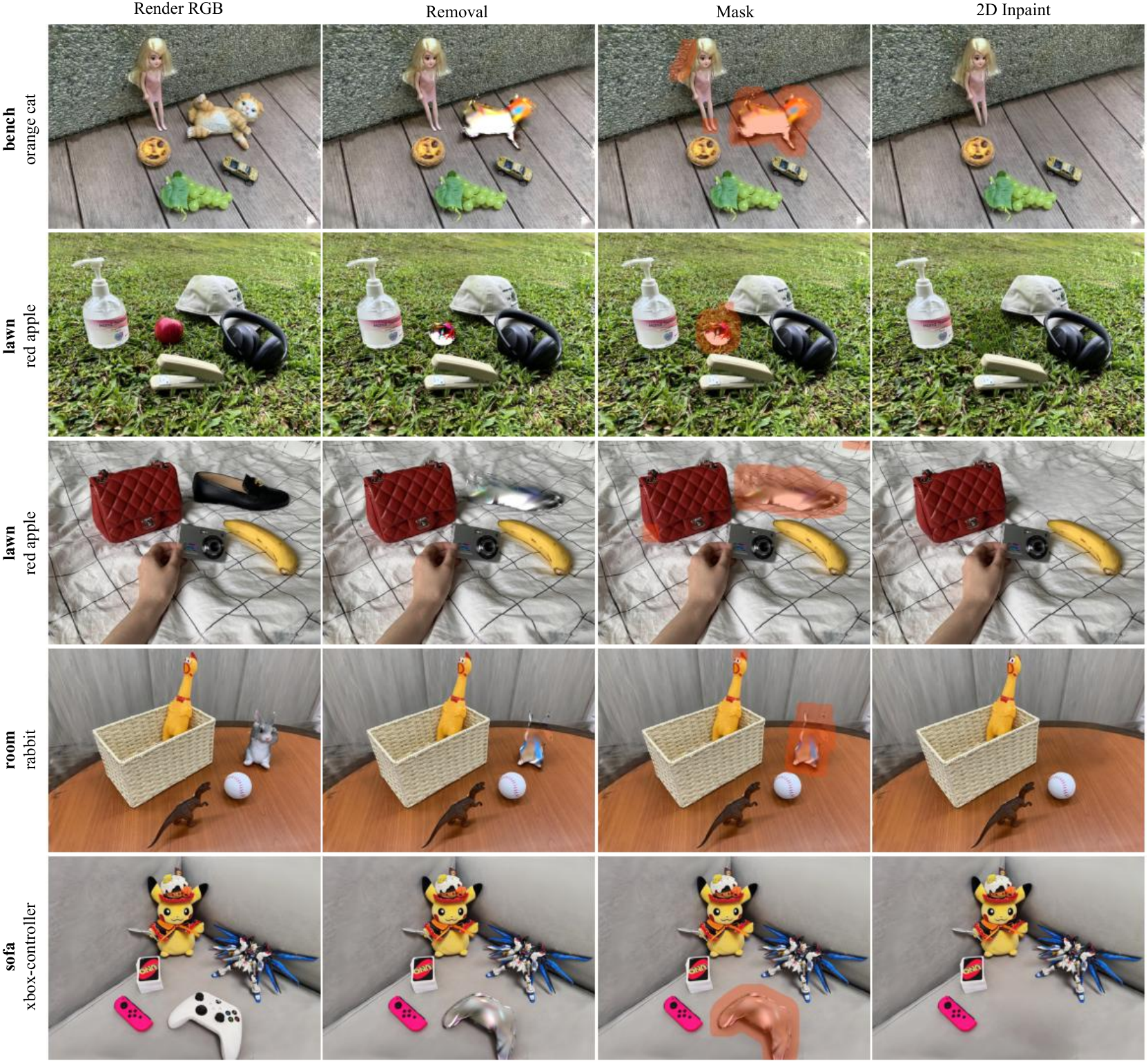}
    \caption{Qualitative visualizations of 3D object removal on 3D-OVS scenes \cite{liu2023weakly3d_ovs}. GaussianDS accurately isolates and prunes target primitives along physical object boundaries, preserving the geometric and appearance fidelity of supporting background surfaces without residual artifacts.}
    \label{fig-objectremoval}
\end{figure}

%--------------------------------------------------------------------------
\subsection{3D Object Removal}
\label{sec:object_removal}
To evaluate whether the learned semantic Gaussian field supports precise downstream spatial scene manipulation, we perform zero-shot 3D object removal on the 3D-OVS benchmark \cite{liu2023weakly3d_ovs, gg_gaussian_grouping}. Specifically, each Gaussian primitive is classified by evaluating the cosine similarity between its latent semantic attribute and the target query embedding or class prototype. Primitives identified as belonging to the queried instance are directly pruned from the 3D Gaussian collection, after which the remaining scene is rendered without modifying the appearance or geometric parameters of unselected Gaussians.

As visualized in Fig.~\ref{fig-objectremoval}, GaussianDS achieves clean object removal across challenging scenes. Because our depth-edge refinement and cross-view mask propagation strictly anchor semantic transitions onto physical geometric boundaries, our method eliminates target primitives with sharp spatial contours. This precision prevents semantic bleeding into adjacent supporting structures, thereby avoiding collateral background erosion while leaving no residual foreground floaters in the edited scene.

%==========================================================================
% SECTION V: CONCLUSION
%==========================================================================
\section{Conclusion}
While 2D foundation models offer rich open-vocabulary semantics, multi-view supervisory contradictions and joint optimization instability frequently cause semantic drift and boundary bleeding in 3D Gaussian representations. GaussianDS resolves this supervision-alignment bottleneck by establishing cross-view mask continuity via pose-aware pseudo-video propagation and co-optimizing appearance, scale-shift-aligned depth, and compact semantics from scratch with depth-edge boundary constraints. Across evaluated benchmarks, GaussianDS improves reconstruction fidelity, open-vocabulary query accuracy, and boundary localization (mBIoU) over multi-stage distillation baselines. Boundary and ablation analyses confirm that these gains stem directly from penalizing semantic leakage along physical depth discontinuities and stabilizing the joint optimization landscape. By replacing cascaded frozen-geometry pipelines with single-stage geometry-grounded co-learning, GaussianDS resolves supervision inconsistency while preserving real-time rendering, providing a reliable foundation for open-vocabulary 3D scene parsing and downstream editing.

\noindent \textbf{Limitations and Future Work.}
While GaussianDS demonstrates robust performance in open-vocabulary 3D scene understanding, its formulation relies primarily on structural depth discontinuities to guide semantic boundaries, which provides weaker constraints across planar, texture-only semantic transitions where geometric depth variation is absent. In future work, we plan to investigate a joint appearance-geometry boundary modeling framework that couples multi-scale photometric contrast with geometric constraints to achieve robust semantic grounding across both physical contours and planar textures.

%==========================================================================
% BIBLIOGRAPHY
%==========================================================================
% \newpage
% \clearpage
\bibliographystyle{ieeetr}
\bibliography{software}

@String(ICCV= {Int. Conf. Comput. Vis.})

@String(ECCV= {Eur. Conf. Comput. Vis.})

@String(TOG= {ACM Trans. Graph.})

@String(AAAI = {AAAI})

@String(ICCV  = {ICCV})

@String(ECCV  = {ECCV})

@String(TOG   = {ACM TOG})

@article{3dgs_kerbl20233d,
  title={3D Gaussian Splatting for Real-Time Radiance Field Rendering},
  author={Kerbl, Bernhard and Kopanas, Georgios and Leimkuehler, Thomas and Drettakis, George},
  journal={ACM Transactions on Graphics (TOG)},
  volume={42},
  number={4},
  pages={1--14},
  year={2023},
  publisher={ACM New York, NY, USA}
}

@inproceedings{depth_anything_yang2024depth,
  title={Depth anything: Unleashing the power of large-scale unlabeled data},
  author={Yang, Lihe and Kang, Bingyi and Huang, Zilong and Xu, Xiaogang and Feng, Jiashi and Zhao, Hengshuang},
  booktitle={Proceedings of the IEEE/CVF Conference on Computer Vision and Pattern Recognition},
  pages={10371--10381},
  year={2024}
}

@inproceedings{clip_radford2021learning,
  title={Learning transferable visual models from natural language supervision},
  author={Radford, Alec and Kim, Jong Wook and Hallacy, Chris and Ramesh, Aditya and Goh, Gabriel and Agarwal, Sandhini and Sastry, Girish and Askell, Amanda and Mishkin, Pamela and Clark, Jack and others},
  booktitle={International conference on machine learning},
  pages={8748--8763},
  year={2021},
  organization={PMLR}
}

@inproceedings{langsplat_qin2024langsplat,
  title={Langsplat: 3d language gaussian splatting},
  author={Qin, Minghan and Li, Wanhua and Zhou, Jiawei and Wang, Haoqian and Pfister, Hanspeter},
  booktitle={Proceedings of the IEEE/CVF Conference on Computer Vision and Pattern Recognition},
  pages={20051--20060},
  year={2024}
}

@inproceedings{depth_r_chung2024depth,
  title={Depth-regularized optimization for 3d gaussian splatting in few-shot images},
  author={Chung, Jaeyoung and Oh, Jeongtaek and Lee, Kyoung Mu},
  booktitle={Proceedings of the IEEE/CVF Conference on Computer Vision and Pattern Recognition},
  pages={811--820},
  year={2024}
}

@InProceedings{sam_Kirillov_2023_ICCV,
    author    = {Kirillov, Alexander and Mintun, Eric and Ravi, Nikhila and Mao, Hanzi and Rolland, Chloe and Gustafson, Laura and Xiao, Tete and Whitehead, Spencer and Berg, Alexander C. and Lo, Wan-Yen and Dollar, Piotr and Girshick, Ross},
    title     = {Segment Anything},
    booktitle = {Proceedings of the IEEE/CVF International Conference on Computer Vision (ICCV)},
    month     = {October},
    year      = {2023},
    pages     = {4015-4026}
}

@InProceedings{dino_Caron_2021_ICCV,
    author    = {Caron, Mathilde and Touvron, Hugo and Misra, Ishan and J\'egou, Herv\'e and Mairal, Julien and Bojanowski, Piotr and Joulin, Armand},
    title     = {Emerging Properties in Self-Supervised Vision Transformers},
    booktitle = {Proceedings of the IEEE/CVF International Conference on Computer Vision (ICCV)},
    month     = {October},
    year      = {2021},
    pages     = {9650-9660}
}

@inproceedings{opengaussian,
    title={OpenGaussian: Towards Point-Level 3D Gaussian-based Open Vocabulary Understanding},
    author={Wu, Yanmin and Meng, Jiarui and Li, Haijie and Wu, Chenming and Shi, Yahao and Cheng, Xinhua and Zhao, Chen and Feng, Haocheng and Ding, Errui and Wang, Jingdong and Zhang, Jian},
    booktitle={Proceedings of the Advances in Neural Information Processing Systems (NeurIPS)},
    pages={19114--19138},
    year={2024}
}

@inproceedings{sugar_guedon2024sugar,
  title={Sugar: Surface-aligned gaussian splatting for efficient 3d mesh reconstruction and high-quality mesh rendering},
  author={Gu{\'e}don, Antoine and Lepetit, Vincent},
  booktitle={Proceedings of the IEEE/CVF Conference on Computer Vision and Pattern Recognition},
  pages={5354--5363},
  year={2024}
}

@inproceedings{wang2024gscream,
  title={Learning 3d geometry and feature consistent gaussian splatting for object removal},
  author={Wang, Yuxin and Wu, Qianyi and Zhang, Guofeng and Xu, Dan},
  booktitle={European conference on computer vision},
  pages={1--17},
  year={2024},
  organization={Springer}
}

@article{chen2023neusg,
  title={Neusg: Neural implicit surface reconstruction with 3d gaussian splatting guidance},
  author={Chen, Hanlin and Li, Chen and Lee, Gim Hee},
  journal={arXiv preprint arXiv:2312.00846},
  year={2023}
}

@article{lyu20243dgsr,
  title={3dgsr: Implicit surface reconstruction with 3d gaussian splatting},
  author={Lyu, Xiaoyang and Sun, Yang-Tian and Huang, Yi-Hua and Wu, Xiuzhe and Yang, Ziyi and Chen, Yilun and Pang, Jiangmiao and Qi, Xiaojuan},
  journal={ACM Transactions on Graphics (ToG)},
  volume={43},
  number={6},
  pages={1--12},
  year={2024},
  publisher={ACM New York, NY, USA}
}

@inproceedings{ravi2024sam2,
  title={Sam 2: Segment anything in images and videos},
  author={Ravi, Nikhila and Gabeur, Valentin and Hu, Yuan-Ting and Hu, Ronghang and Ryali, Chaitanya and Ma, Tengyu and Khedr, Haitham and R{\"a}dle, Roman and Rolland, Chloe and Gustafson, Laura and others},
  booktitle={International Conference on Learning Representations},
  volume={2025},
  pages={28085--28128},
  year={2025}
}

@inproceedings{gg_gaussian_grouping,
    title={Gaussian Grouping: Segment and Edit Anything in 3D Scenes},
    author={Ye, Mingqiao and Danelljan, Martin and Yu, Fisher and Ke, Lei},
    booktitle={ECCV},
    year={2024}
}

@inproceedings{shi2024languageembedding,
  title={Language embedded 3d gaussians for open-vocabulary scene understanding},
  author={Shi, Jin-Chuan and Wang, Miao and Duan, Hao-Bin and Guan, Shao-Hua},
  booktitle={Proceedings of the IEEE/CVF Conference on Computer Vision and Pattern Recognition},
  pages={5333--5343},
  year={2024}
}

@article{kerbl2024hierarchical,
  title={A hierarchical 3d gaussian representation for real-time rendering of very large datasets},
  author={Kerbl, Bernhard and Meuleman, Andreas and Kopanas, Georgios and Wimmer, Michael and Lanvin, Alexandre and Drettakis, George},
  journal={ACM Transactions on Graphics (TOG)},
  volume={43},
  number={4},
  pages={1--15},
  year={2024},
  publisher={ACM New York, NY, USA}
}

@inproceedings{jin20243dfires,
  title={3DFIRES: Few Image 3D REconstruction for Scenes with Hidden Surfaces},
  author={Jin, Linyi and Kulkarni, Nilesh and Fouhey, David F},
  booktitle={Proceedings of the IEEE/CVF Conference on Computer Vision and Pattern Recognition},
  pages={9742--9751},
  year={2024}
}

@inproceedings{zhou2024feature,
  title={Feature 3dgs: Supercharging 3d gaussian splatting to enable distilled feature fields},
  author={Zhou, Shijie and Chang, Haoran and Jiang, Sicheng and Fan, Zhiwen and Zhu, Zehao and Xu, Dejia and Chari, Pradyumna and You, Suya and Wang, Zhangyang and Kadambi, Achuta},
  booktitle={Proceedings of the IEEE/CVF Conference on Computer Vision and Pattern Recognition},
  pages={21676--21685},
  year={2024}
}

@inproceedings{liao2024clipgs,
  title={CLIP-GS: Unifying vision-language representation with 3D Gaussian splatting},
  author={Jiao, Siyu and Dong, Haoye and Yin, Yuyang and Jie, Zequn and Qian, Yinlong and Zhao, Yao and Shi, Humphrey and Wei, Yunchao},
  booktitle={2025 IEEE/CVF International Conference on Computer Vision (ICCV)},
  pages={4670--4680},
  year={2025},
  organization={IEEE}
}

@inproceedings{lsegli2022languagedriven,
title={Language-driven Semantic Segmentation},
author={Boyi Li and Kilian Q Weinberger and Serge Belongie and Vladlen Koltun and Rene Ranftl},
booktitle={International Conference on Learning Representations},
year={2022},
url={https://openreview.net/forum?id=RriDjddCLN}
}

@inproceedings{lai2024lisa,
  title={Lisa: Reasoning segmentation via large language model},
  author={Lai, Xin and Tian, Zhuotao and Chen, Yukang and Li, Yanwei and Yuan, Yuhui and Liu, Shu and Jia, Jiaya},
  booktitle={Proceedings of the IEEE/CVF Conference on Computer Vision and Pattern Recognition},
  pages={9579--9589},
  year={2024}
}

@inproceedings{lerf2023,
 author = {Kerr, Justin and Kim, Chung Min and Goldberg, Ken and Kanazawa, Angjoo and Tancik, Matthew},
 title = {LERF: Language Embedded Radiance Fields},
 booktitle = {International Conference on Computer Vision (ICCV)},
 year = {2023},
}

@article{liu2023weakly3d_ovs,
  title={Weakly supervised 3d open-vocabulary segmentation},
  author={Liu, Kunhao and Zhan, Fangneng and Zhang, Jiahui and Xu, Muyu and Yu, Yingchen and El Saddik, Abdulmotaleb and Theobalt, Christian and Xing, Eric and Lu, Shijian},
  journal={Advances in Neural Information Processing Systems},
  volume={36},
  pages={53433--53456},
  year={2023}
}

@article{bhat2023zoedepth,
  title={Zoedepth: Zero-shot transfer by combining relative and metric depth},
  author={Bhat, Shariq Farooq and Birkl, Reiner and Wofk, Diana and Wonka, Peter and M{\"u}ller, Matthias},
  journal={arXiv preprint arXiv:2302.12288},
  year={2023}
}

@inproceedings{yu2024cogs,
  title={Cogs: Controllable gaussian splatting},
  author={Yu, Heng and Julin, Joel and Milacski, Zolt{\'a}n {\'A} and Niinuma, Koichiro and Jeni, L{\'a}szl{\'o} A},
  booktitle={Proceedings of the IEEE/CVF Conference on Computer Vision and Pattern Recognition},
  pages={21624--21633},
  year={2024}
}

@inproceedings{yin2024sai3d,
  title={Sai3d: Segment any instance in 3d scenes},
  author={Yin, Yingda and Liu, Yuzheng and Xiao, Yang and Cohen-Or, Daniel and Huang, Jingwei and Chen, Baoquan},
  booktitle={Proceedings of the IEEE/CVF Conference on Computer Vision and Pattern Recognition},
  pages={3292--3302},
  year={2024}
}

@article{oquab2023dinov2,
  title={Dinov2: Learning robust visual features without supervision},
  author={Oquab, Maxime and Darcet, Timoth{\'e}e and Moutakanni, Th{\'e}o and Vo, Huy and Szafraniec, Marc and Khalidov, Vasil and Fernandez, Pierre and Haziza, Daniel and Massa, Francisco and El-Nouby, Alaaeldin and others},
  journal={arXiv preprint arXiv:2304.07193},
  year={2023}
}

@article{hu2024sparselgs,
  title={Sparselgs: Sparse view language embedded gaussian splatting},
  author={Hu, Jun and Chen, Zhang and Li, Zhong and Xu, Yi and Zhang, Juyong},
  journal={IEEE Transactions on Visualization and Computer Graphics},
  year={2026},
  publisher={IEEE}
}

@misc{ren2024groundedsam,
      title={Grounded SAM: Assembling Open-World Models for Diverse Visual Tasks}, 
      author={Tianhe Ren and Shilong Liu and Ailing Zeng and Jing Lin and Kunchang Li and He Cao and Jiayu Chen and Xinyu Huang and Yukang Chen and Feng Yan and Zhaoyang Zeng and Hao Zhang and Feng Li and Jie Yang and Hongyang Li and Qing Jiang and Lei Zhang},
      year={2024},
      eprint={2401.14159},
      archivePrefix={arXiv},
      primaryClass={cs.CV}
}

@inproceedings{dngaussian,
  title={DNGaussian: Optimizing sparse-view 3D Gaussian radiance fields with global-local depth normalization},
  author={Li, Jiahe and Zhang, Jiawei and Bai, Xiao and Zheng, Jin and Ning, Xin and Zhou, Jun and Gu, Lin},
  booktitle={Proceedings of the IEEE/CVF Conference on Computer Vision and Pattern Recognition},
  pages={20775--20785},
  year={2024}
}

@inproceedings{cen2025segment,
  title={Segment Any 3D Gaussians},
  author={Cen, Jiazhong and Fang, Jiemin and Yang, Chen and Xie, Lingxi and Zhang, Xiaopeng and Tian, Qi},
  booktitle={Proceedings of the AAAI Conference on Artificial Intelligence},
  volume={39},
  number={2},
  pages={1971--1979},
  year={2025}
}

@inproceedings{tian2025ccllgs,
  title={CCL-LGS: Contrastive Codebook Learning for 3D Language Gaussian Splatting},
  author={Tian, Linrui and Li, Xin and Ma, Li and Zhang, Xiaopeng and Tian, Qi},
  booktitle={Proceedings of the IEEE/CVF International Conference on Computer Vision},
  pages={9855--9864},
  year={2025}
}

@inproceedings{zhu2025objectgs,
  title={ObjectGS: Object-level 3D Gaussian Splatting for Scene Decomposition and Editing},
  author={Zhu, Bangyan and Lyu, Weijie and Zhou, Ziyue and Huang, Xinyue and Zhang, Xiaopeng},
  booktitle={Proceedings of the IEEE/CVF International Conference on Computer Vision},
  year={2025}
}

@article{li2025langsplatv2,
  title={Langsplatv2: High-dimensional 3d language gaussian splatting with 450+ fps},
  author={Li, Wanhua and Zhao, Yujie and Qin, Minghan and Liu, Yang and Cai, Yuanhao and Gan, Chuang and Pfister, Hanspeter},
  journal={Advances in Neural Information Processing Systems},
  volume={38},
  pages={174306--174330},
  year={2026}
}

@inproceedings{peng2026gags,
  title={Gags: Granularity-aware feature distillation for language gaussian splatting},
  author={Peng, Yuning and Wang, Haiping and Liu, Yuan and Wen, Chenglu and Dong, Zhen and Yang, Bisheng},
  booktitle={Proceedings of the AAAI Conference on Artificial Intelligence},
  volume={40},
  number={10},
  pages={8376--8384},
  year={2026}
}

@inproceedings{qu2024goi,
  title={GOI: Find 3D Gaussians of Interest with an Optimizable Open-vocabulary Semantic-space Hyperplane},
  author={Qu, Yansong and Dai, Shaohui and Li, Xinyang and Lin, Jianghang and Cao, Liujuan and Zhang, Shengchuan and Ji, Rongrong},
  booktitle={Proceedings of the 32nd ACM International Conference on Multimedia},
  pages={5328--5337},
  year={2024}
}

@article{SUN2026105830,
title = {CAGS: Open-vocabulary 3D scene understanding with context-aware Gaussian splatting},
journal = {Image and Vision Computing},
volume = {165},
pages = {105830},
year = {2026},
issn = {0262-8856},
doi = {https://doi.org/10.1016/j.imavis.2025.105830},
url = {https://www.sciencedirect.com/science/article/pii/S0262885625004184},
author = {Wei Sun and Yuan Li and Jianbin Jiao}
}

@article{he2025mpd,
  title={MPD-GS: Mask-guided point densification for Gaussian splatting},
  author={He, Junhui and Xiao, Wen and Wang, Guilong and Cheng, Jiteng and Zhang, Jiaxing and Yang, Chao},
  journal={Neurocomputing},
  pages={132438},
  year={2025},
  publisher={Elsevier}
}

@article{zhao2026sdp,
  title={SDP-GS: Sparse-view Gaussian splatting via segmentation-aware depth priors},
  author={Zhao, Qi and Deng, Yangyan and Zhang, Jiawei and Zhang, Hong and Yang, Yifan and Yuan, Ding},
  journal={Neurocomputing},
  pages={133831},
  year={2026},
  publisher={Elsevier}
}

@article{zhang2026dgc,
  title={DGC-GS: Enhancing geometric consistency in sparse-view 3D Gaussian splatting},
  author={Zhang, Zhitao and Wang, Ping and Zhang, Zhanhua and Cao, Xingjian and Tian, Can and Yang, Shuo and Ding, Xiaobing and Ge, Wancheng},
  journal={Neurocomputing},
  pages={134623},
  year={2026},
  publisher={Elsevier}
}

@article{wang20263d,
  title={A 3D Gaussian reconstruction method with monocular depth optimization and dynamic KNN densification},
  author={Wang, Jinhua and Li, Yawei and Cao, Jie},
  journal={Neurocomputing},
  pages={133355},
  year={2026},
  publisher={Elsevier}
}

%%
%% End of file
\end{document}